\documentclass{sig-alternate-05-2015}
\pdfoutput=1

\usepackage{times}
\usepackage{graphicx}
\graphicspath{{figures/}}
\usepackage{subfigure}
\usepackage{algorithm}
\usepackage{algorithmic}

\usepackage{ctable}
\usepackage{commath}
\usepackage{balance}
\usepackage{setspace}
\usepackage{microtype}
\usepackage{balance}

\newcommand{\argmin}{\operatornamewithlimits{arg\,min}}

\newcommand{\eg}{{\it e.g.}}
\newcommand{\ie}{{\it i.e.}}
\newcommand{\etal}{{\it et al.}}
\newcommand{\figurename}{Figure}
\newcommand{\tool}{deepTarget}

\newcommand{\mytablespacing}{1}
\newcommand{\bhlcolor}{black}
\newcommand{\finalcolor}{black}

\begin{document}
\CopyrightYear{2016}
\setcopyright{acmcopyright}
\conferenceinfo{BCB '16,}{October 02-05, 2016, Seattle, WA, USA}
\isbn{978-1-4503-4225-4/16/10}\acmPrice{\$15.00}
\doi{http://dx.doi.org/10.1145/2975167.2975212}

\title{\tool: End-to-end Learning Framework for microRNA Target Prediction using Deep Recurrent Neural Networks}

\numberofauthors{4}
\author{
\alignauthor
Byunghan Lee\\
    \affaddr{Electrical and Computer Eng.}\\
    \affaddr{Seoul National University}\\
    \affaddr{Seoul 08826, Korea}
\alignauthor
Junghwan Baek\\
    \affaddr{Interdisciplinary Program in Bioinformatics}\\
    \affaddr{Seoul National University}\\
    \affaddr{Seoul 08826, Korea}
\and
\alignauthor
Seunghyun Park\\
    \affaddr{Electrical and Computer Eng.}\\
    \affaddr{Seoul National University}\\
    \affaddr{Seoul 08826, Korea}\\
    \affaddr{Electrical Engineering}\\
    \affaddr{Korea University}\\
    \affaddr{Seoul 02841, Korea}
%\and
\alignauthor
Sungroh Yoon\titlenote{To whom correspondence should be addressed.}\\
    \affaddr{Electrical and Computer Eng. \& Interdisciplinary Program in Bioinformatics}\\
    \affaddr{Seoul National University}\\
    \affaddr{Seoul 08826, Korea}\\
%    \affaddr{Neurology \& Neurological Sciences}\\
%    \affaddr{Stanford University}\\
%    \affaddr{Stanford, CA94305, USA}\\
\email{sryoon@snu.ac.kr}}

\maketitle

\begin{abstract}
%2000 char max
MicroRNAs (miRNAs) are short sequences of ribonucleic acids that control the expression of target messenger RNAs (mRNAs) by binding them. Robust prediction of miRNA-mRNA pairs is of utmost importance in deciphering \textcolor{\bhlcolor}{gene regulation} but has been challenging because of high false positive rates, despite a deluge of computational tools that normally require laborious manual feature extraction.
This paper presents an end-to-end machine learning framework for miRNA target prediction. Leveraged by deep recurrent neural networks-based auto-encoding and sequence-sequence interaction learning, our approach not only delivers an unprecedented level of accuracy but also eliminates the need for manual feature extraction. The performance gap between the proposed method and existing alternatives is substantial (over 25\% increase in F-measure), and \tool~delivers a quantum leap in the long-standing challenge of robust miRNA target prediction. \textcolor{\bhlcolor}{[availability: http://data.snu.ac.kr/pub/deepTarget ]}

\iffalse
MicroRNAs (miRNAs) are short sequences of ribonucleic acids that control the expression of target messenger RNAs (mRNAs) by binding them. Robust prediction of miRNA-mRNA pairs is of utmost importance in deciphering gene regulation but has been challenging because of high false positive rates, despite a deluge of computational tools that normally require laborious manual feature extraction. This paper presents an end-to-end machine learning framework for the miRNA target prediction. Leveraged by deep recurrent neural networks-based auto-encoding and sequence-sequence interaction learning, our approach not only delivers an unprecedented level of accuracy but also eliminates the need for manual feature extraction. The performance gap between the proposed method and existing alternatives is substantial (over 75% increase in F-measure), and deepTarget delivers a quantum leap in the long-standing challenge of robust miRNA target prediction.
\fi 
\end{abstract}




\section{Introduction}
\begin{figure}[b!]
    \centering
    \includegraphics[width=\linewidth]{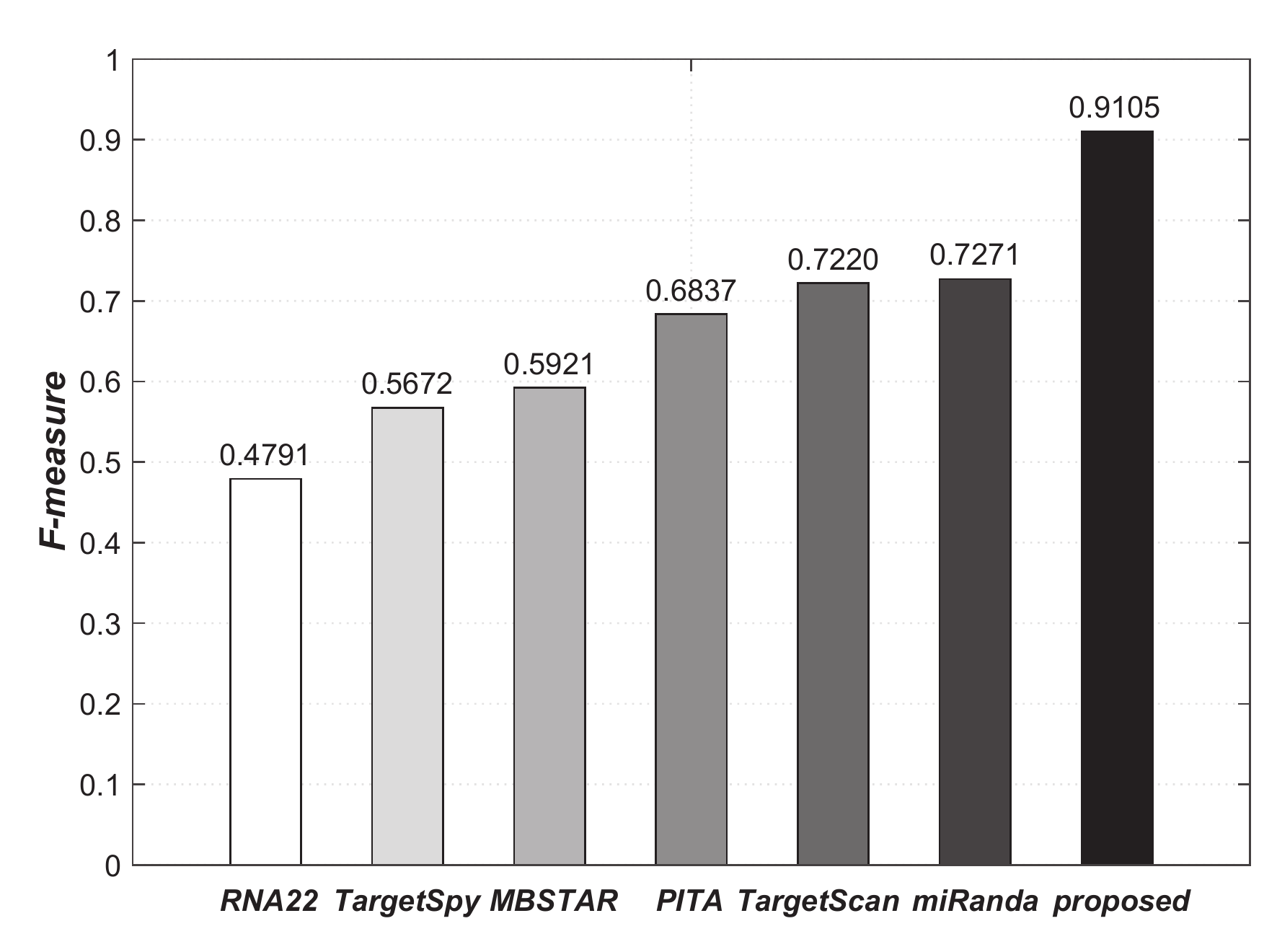}
    \caption{F-measure comparison of miRNA target prediction tools: The proposed \tool~method significantly outperforms the alternatives  (see Section 4 for details)}
    \label{fig:intro-proposed}
\end{figure}

\begin{figure*}
    \centering
    \includegraphics[width=.9\linewidth]{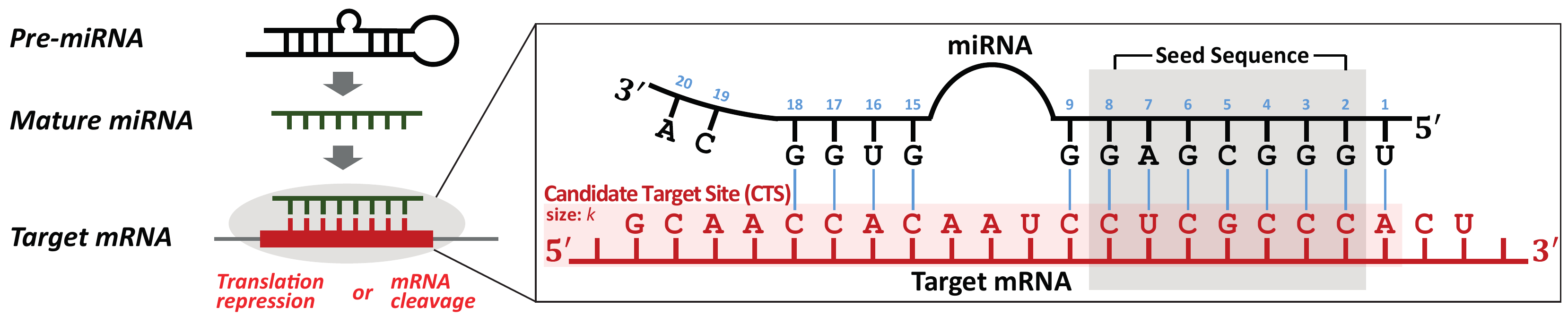}
    \caption{miRNA biogenesis and its function: microRNAs (miRNAs) exhibit their regulatory function by binding to the target sites present in the 3' untranslated region (UTR) of their cognate mRNAs. The \emph{seed} sequence of a miRNA is defined as the first two to eight nucleotides starting from the 5' to the 3' UTR. The \textit{candidate target site} (CTS) refers to the small segment of length $k$ of mRNA. The blue line between a pair of nucleotides (one from miRNA and the other from mRNA) represents the Watson-Crick (WC) pairing formed by sequence complementarity (\ie, the binding of $\texttt{A}$-$\texttt{U}$ and that of $\texttt{C}$-$\texttt{G}$)}
    \label{fig:intro-mirna}
\end{figure*}

MicroRNAs (miRNAs) are small non-coding RNA molecules that can control the function of their target messenger RNAs (mRNAs) by down-regulating the expression of the targets~\cite{bartel2004micrornas}. By controlling the gene expression at the RNA level, miRNAs are known to be involved in various biological processes and diseases~\cite{lu2005microrna}. As miRNAs play a central role in the post-transcriptional regulation of more than 60\% of protein coding genes~\cite{friedman2009most}, investigating miRNAs is of utmost importance in many disciplines of life science. %As informally shown in \figurename~\ref{fig:intro-mirna}, 
	As explained further in Section 2.3, miRNAs are derived from the precursor miRNAs (pre-miRNAs) and then exhibit their regulatory function by binding to the target sites present in mRNAs. Two types of computational problems about miRNAs thus naturally arise in bioinformatics: miRNA host identification (\ie, the problem of locating the genes that encode pre-miRNAs) and miRNA target prediction (\ie, the task of finding the mRNA targets of given miRNAs). This paper focuses on the target prediction problem.

Due to the importance, there has been a deluge of computational algorithms proposed to address the miRNA target prediction problem~\cite{min2010got,peterson2014common,yoon2006computational}. However, most of the existing approaches suffer from two major limitations, often failing to deliver satisfactory performance in practice (see \figurename~\ref{fig:intro-proposed}). First, the conventional approaches make predictions based on the manually crafted features of known miRNA-mRNA pairs [\eg, the level of sequence complementarity (see Section~\ref{sec:background-def}) between a miRNA sequence and the sequence of a binding site in its target mRNA]. Manual feature extraction is time-consuming, laborious, and error-prone, giving no guarantee for generality. Second, existing tools often fail to effectively filter out false positives [\ie, bogus miRNA-mRNA pairs that do not actually interact \emph{in vivo}], producing an unacceptable \textcolor{\bhlcolor}{low level of specificity}. This challenge originates from the fact that there are only four types of letters (\texttt{A}, \texttt{C}, \texttt{G}, and \texttt{U}) in RNA sequences and that the length of a typical miRNA sequence is short (about 22 nucleotides), producing a high chance of seeing a random site (in mRNA) whose sequence is complementary to the miRNA.

To boost the sensitivity of miRNA target prediction, a variety of features have been proposed. According to Menor et al.~\cite{menor2014mirmark}, as many as 151 kinds of features appear in the literature, which can be broadly grouped into four common types~\cite{peterson2014common}: the degree of Watson-Crick matches of a seed sequence (see \figurename~\ref{fig:intro-mirna}); the degree of sequence conservation across species; Gibbs free energy, which measures the stability of the binding of a miRNA-mRNA pair, and the site accessibility, which measures the hybridization possibility of a pair from their secondary structures.

%Among these laboriously extracted features, the first order target binding signal is the sequence complementarity between miRNAs and mRNAs. However, owing to the fact that miRNAs are about 22 nucleotides long while mRNAs are much longer than that, the number of false positive target sites is high.

To address the limitations of existing approaches, this paper proposes \tool, an end-to-end machine learning framework for robust miRNA target prediction. \tool~adopts deep recurrent neural network (RNN)-based auto-encoders to discover the inherent representations of miRNA and mRNA sequences and utilizes stacked RNNs to learn the sequence-to-sequence interactions underlying miRNA-mRNA pairs. Leveraged by this combination of unsupervised and supervised learning approaches, \tool~not only delivers an unprecedented level of accuracy but also eliminates the need for manual feature extraction. Furthermore, according to our visual inspection of the activation of the RNN layers used in \tool, meaningful patterns appeared in the nucleotide positions that corresponded to the real miRNA-mRNA binding sites.  Further analyzing the activation patterns may allow us to obtain novel biological insight into regulatory interactions. %in an automated way. % without any sequence alignment that often limits the robustness of target prediction.

 %To facilitate biological understanding of the factors that discriminate positive and negative examples, \tool~also provides a visualization technique the activation in the RNN layers

As shown in \figurename~\ref{fig:intro-proposed}, the performance gap between \tool~and the compared alternatives is substantial (over 25\% increase in F-measure), and \tool~delivers a quantum leap in the long-standing challenge of robust miRNA target prediction.

%The rest of the paper is organized as follows. Section 2 presents the background of the building blocks of the proposed approach. In Section 3, we describe the proposed approach that relies on autoencoder and recurrent neural networks. The experimental results are presented in Section 4. 

\section{Background}
\begin{figure*}
    \centering
    \includegraphics[width=0.5\linewidth]{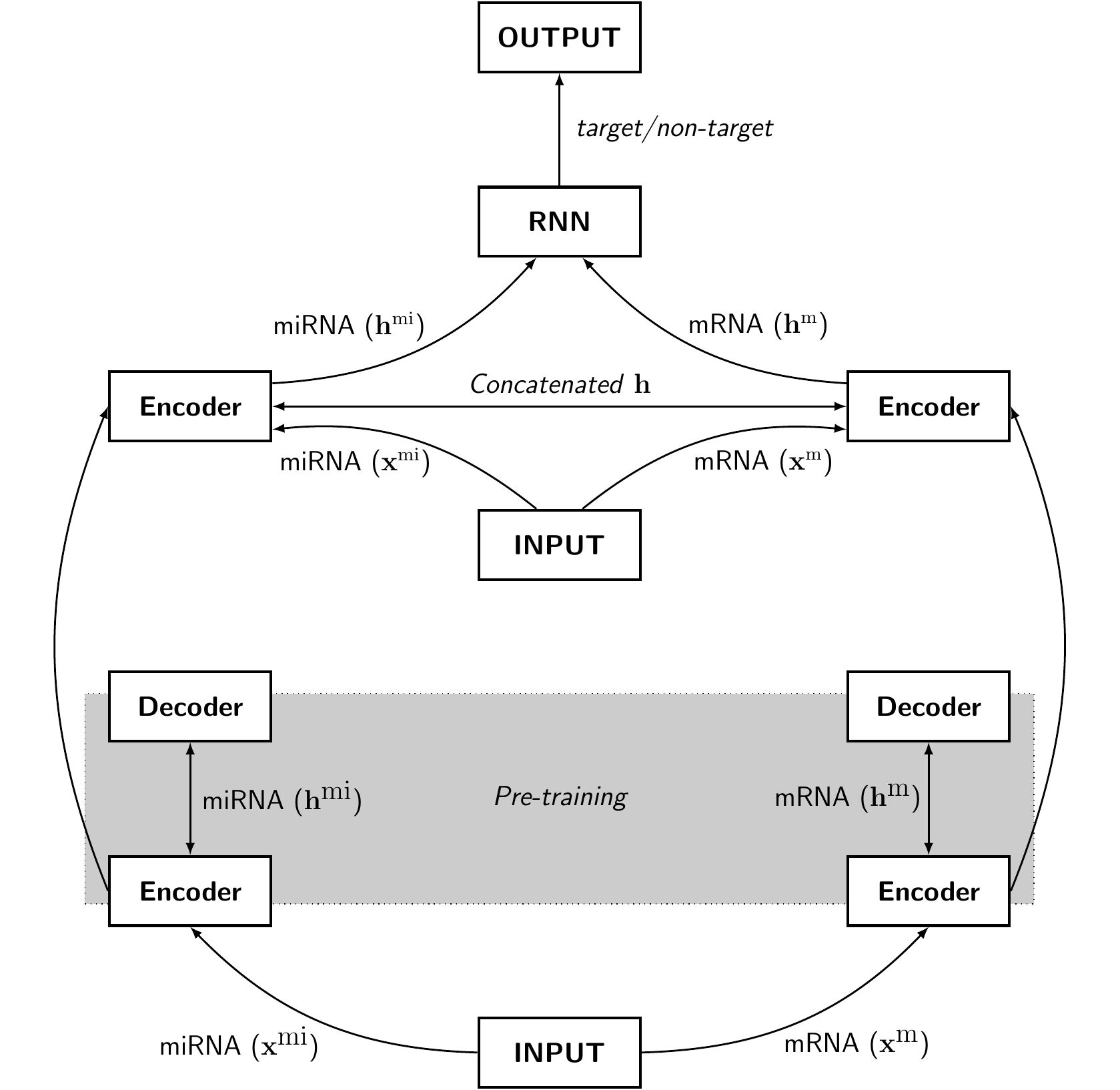}
    \caption{Overview of proposed \tool~methodology: Our method is based on RNA sequence modeling and sequence-to-sequence interaction learning using autoencoders and stacked RNNs. The input layer is connected to the first layer of two autoencoders to model miRNA and mRNA sequences, respectively. The second layer is an RNN layer to model the interaction between miRNA and mRNA sequences. The outputs of the top RNN layer are fed into a fully connected output layer, which contains two units for classifying targets and non-targets
}
    \label{fig:methods-overview}
\end{figure*}

As mentioned in the previous section, miRNAs play a crucial role in controlling the function of their target genes by down-regulating gene expression, actively participating in various biological processes.
To predict targets of given miRNAs, numerous computational tools have been proposed~\cite{peterson2014common,srivastava2014comparison}.

\textcolor{\bhlcolor}{There exists a recent study~\cite{chengmirtdl} that used convolutional neural network (CNN) with 20 manually crafted features to predict the target of a miRNA. In \cite{chengmirtdl}, however, the deep learning method was used as a classifier rather than a feature learner from the raw features (\ie, biological sequences)}. The novelty of our approach comes from its use of deep RNNs \textcolor{\bhlcolor}{without laborious feature
engineering}: \tool~performs RNA sequence modeling using autoencoders and learns the interactions between miRNAs and mRNAs using stacked RNNs.
To the best of our knowledge, the proposed methodology is one of the first attempts to predict miRNA targets by sequence modeling using deep RNNs.

%In the following sections, we briefly describe what the autoencoder and recurrent neural network are.

\subsection{Autoencoder}
An autoencoder is an artificial neural network used for learning representations.
Autoencoders have almost similar structure to that of multilayer perceptrons (MLPs) except that the output layer has the same number of nodes as the input layer.
The objective of an autoencoder is to learn a meaningful encoding for a set of data in an unsupervised manner while a MLP is to be trained to predict the target value in a supervised manner.

Typically, an autoencoder consists of two parts, an encoder and a decoder, which can be defined as transitions $\phi$ and $\psi$, respectively, given by
\begin{align}
	\phi: \mathcal{S} \mapsto \mathcal{X} \quad\text{and}\quad \psi: \mathcal{X} \mapsto \mathcal{S}
\end{align}
subject to
\begin{align}
	\argmin_{\phi, \psi}\lVert S - (\psi \circ \phi)S \rVert^{2}
\end{align}
\textcolor{\finalcolor}{for $S \in \mathcal{S}$}, where $\mathcal{S}$ is the input space and $\mathcal{X}$ is the mapped feature space.

Various techniques exist to improve the generalization performance of autoencoders. Examples include the denoising autoencoder (DAE)~\cite{vincent2008extracting}, and the variational autoencoder (VAE)~\cite{kingma2013auto}.
Recently, the VAE concept has become more widely used for learning generative models of data~\cite{gregor2015draw}.

In this paper, RNN-based autoencoders are used to model miRNA and mRNA sequences to extract their inherent features in an unsupervised manner.

\subsection{Recurrent Neural Network (RNN)}
An RNN is an artificial neural network where connections between units form a directed cycle. This creates an internal state of the network that allows it to exhibit dynamic temporal behavior. RNNs can use their internal memory to process sequences of inputs. This makes RNNs applicable to various tasks, such as unsegmented connected handwriting recognition~\cite{le2015simple} and speech recognition~\cite{sak2014long}.

Leveraged by advances in the memory-based units [\eg, long short-term memory (LSTM)~\cite{hochreiter1997long} and gated recurrent unit (GRU)~\cite{cho2014properties}], network architecture [\eg, stacked RNNs~\cite{pascanu2013construct} and bidirectional RNNs~\cite{schuster1997bidirectional}] and training methods [\eg, iRNN~\cite{le2015simple} and dropout~\cite{srivastava2014dropout}], RNNs are delivering breakthrough performance in tasks involving sequential modeling and prediction~\cite{bahdanau2014neural,cho2014learning,sak2014long,vinyals2014show}.

%According to \cite{Goodfellow-et-al-2016-Book}, RNNs can do various sequence modeling tasks: (1) mapping an input sequence to a fixed-size prediction, (2) modeling a distribution over sequences and generating new ones from the estimated distribution, (3) mapping a fixed-length vector into a distribution over sequences, (4) mapping a variable-length input sequence to an output sequence of the same length, and (5) mapping an input sequence to an output sequence that is not necessarily of the same length.

According to Goodfellow et al.~\cite{Goodfellow-et-al-2016-Book}, RNNs can do various sequence modeling tasks including the following:
\begin{itemize}
	\item [\textbf{T1:}] mapping an input sequence to a fixed-size prediction
	\item [\textbf{T2:}] modeling a distribution over sequences and generating new ones from the estimated distribution
	\item [\textbf{T3:}] mapping a fixed-length vector into a distribution over sequences
	\item [\textbf{T4:}] mapping a variable-length input sequence to an output sequence of the same length
	\item [\textbf{T5:}] mapping an input sequence to an output sequence that is not necessarily of the same length
\end{itemize}

In this paper, RNNs are used to perform task \textbf{T5} for sequence modeling using autoencoders and task \textbf{T1} for target prediction. %That is, \tool~utilizes RNNs as the building blocks of autoencoders and models interaction between RNAs.

\subsection{Biology of miRNA-mRNA Interactions}\label{sec:background-def}
In molecular biology, directionality of a nucleic acid, 5'-end and 3'-end, is the end-to-end chemical orientation of a single strand of the nucleic acid. The 3' untranslated region (UTR) of mRNA is the region that directly follows the translation termination codon.
As shown in \figurename~\ref{fig:intro-mirna}, miRNAs exhibit their function by binding to the target sites present in the 3' UTR of their cognate mRNAs.
By binding to target sites within the 3' UTR, miRNAs can decrease gene expression of mRNAs by either inhibiting translation or causing degradation of the transcript.
The \textit{seed} sequence of a miRNA is defined as the first two to eight nucleotides starting from the 5' to the 3' UTR. The \textit{candidate target site} (CTS) refers to the small segment of length $k$ of mRNA that contains \textcolor{\bhlcolor}{a complementary match to the} seed region at the head. The blue line between a pair of nucleotides (one from miRNA and the other from mRNA) in \figurename~\ref{fig:intro-mirna} represents the Watson-Crick (WC) pairing formed by sequence complementarity (\ie, the binding of $\texttt{A}$-$\texttt{U}$ and that of $\texttt{C}$-$\texttt{G}$).

The canonical sites are complementary to the miRNA seed region while the non-canonical sites are with bulges or single-nucleotide loops in the seed region. The non-canonical sites include two types, 3' compensatory and centered sites. The proposed machine learning-based approach has potential to learn both canonical and non-canonical sites.
While considering CTSs to predict targets of given miRNAs, there are four types of seed sequence matches~\cite{peterson2014common} often referred to as 6mer, 7mer-m8, 7mer-A1, and 8mer in the literature. The details of each type are as follows:
\begin{itemize}
	\item 6mer: exact WC pairing between the miRNA seed and mRNA for six nucleotides
	\item 7mer-m8: exact WC pairing from nucleotides 2--8 of the miRNA seed
	\item 7mer-A1: exact WC pairing from nucleotides 2--7 of the miRNA seed in addition to an \texttt{A} of the miRNA nucleotide 1
	\item 8mer: exact WC pairing from nucleotides 2--8 of the miRNA seed in addition to an \texttt{A} of the miRNA nucleotide 1
\end{itemize}

%(1) 6mer: exact WC pairing between the miRNA seed and mRNA for six nucleotides; (2) 7mer-m8: exact WC pairing from nucleotides 2--8 of the miRNA seed; (3) 7mer-A1: exact WC pairing from nucleotides 2--7 of the miRNA seed in addition to an \texttt{A} of the miRNA nucleotide 1; and (4) 8mer: exact WC pairing from nucleotides 2--8 of the miRNA seed in addition to an \texttt{A} of the miRNA nucleotide 1. 

\section{Proposed Methodology}
\renewcommand{\algorithmiccomment}[1]{\hfill \(\triangleright\)~#1}

\begin{algorithm*}[tb]
    \caption{Pseudo-code of \tool}
    \label{alg:methods-rnn}
\begin{algorithmic}[1]
    \STATE {\bfseries Input:} $N$ encoded RNA sequences, $\mathbf{x}_{1}^{\text{mi}}, \cdots, \mathbf{x}_{N}^{\text{mi}}$ and $\mathbf{x}_{1}^{\text{m}}, \cdots, \mathbf{x}_{N}^{\text{m}}$
    \STATE {\bfseries Output:} $\mathbf{y}$ (target/non-target)

    \STATE Pre-train the autoencoders $\text{AE}^{\text{mi}}$ and $\text{AE}^{\text{m}}$
    \COMMENT {$\text{AE}^{\text{mi}}$, $\text{AE}^{\text{m}}$: the autoencoders described in Section~\ref{sec:methods-ae}}
    \REPEAT
    \STATE minimize the reconstruction error~
    $\mathcal{L}(\mathbf{x}, \mathbf{\hat{x}}) = \lVert \mathbf{x} - \mathbf{\hat{x}} \rVert^{2}$
    \UNTIL{\# of epoch is $n_{\text{epoch}}$}

    \STATE Define the whole architecture [$\text{EN}^{\text{mi}} \parallel \text{EN}^{\text{m}}$]-RNN
    \COMMENT {$\text{EN}^{\text{mi}}$, $\text{EN}^{\text{m}}$: the encoders in described in Section~\ref{sec:methods-rnn}}
    \STATE Concatenate two representations, $\mathbf{h}^{\text{mi}}$ and $\mathbf{h}^{\text{m}}$, into $\mathbf{h}$
    \COMMENT {$\mathbf{h}^{\text{mi}}$, $\mathbf{h}^{\text{m}}$: the outputs of each encoder}
    \STATE Fine-tune the architecture
    \REPEAT
    \STATE minimize the logarithmic loss~
    $\mathcal{L}(\mathbf{w}) = -\frac{1}{N} \sum_{i=1}^{N} (p_{i}\textrm{log}(p_{i}) + (1-p_{i})\textrm{log}(1-p_{i}))$
    \UNTIL{\# of epoch is $n_{\text{epoch}}$}
\end{algorithmic}
\end{algorithm*}

\figurename~\ref{fig:methods-overview} shows the overview of the proposed \tool~methodology. Our method is based on RNA sequence modeling and sequence-to-sequence interaction learning using autoencoders and stacked RNNs.
The input layer is connected to the first layer of two autoencoders to model miRNA and mRNA sequences, respectively. The second layer is an RNN layer to model the interaction between miRNA and mRNA sequences. The outputs of the top RNN layer are fed into a fully connected output layer, which contains two units for classifying targets and non-targets.

We preprocess the CTSs with length $k$, which are empirically determined by all of the four seed matching types between miRNAs and mRNAs as described in Section~\ref{sec:background-def}. After preprocessing, as shown in Algorithm~\ref{alg:methods-rnn}, our approach proceeds in two major steps with miRNA and CTS sequence pair as an input: (1) unsupervised learning of two autoencoders for modeling miRNAs and mRNAs (lines 4--7), and (2) supervised learning of the whole architecture for modeling interaction between miRNAs and mRNAs (lines 10--12).

Throughout this paper, we denote a sequence as $\mathbf{s}^{\text{mi}}$ of $n_{d}^{\text{mi}}$-dimensional vector for miRNA and $\mathbf{s}^{\text{m}}$ of $n_{d}^{\text{m}}$-dimensional vector for mRNA where $n_{d}^{\text{mi}}$ and $n_{d}^{\text{m}}$ can, in general, be different and $k \approx \text{max}(n_{d}^{\text{mi}})$.

\subsection{Input Representation}
Each RNA read is a sequence that has four types of nucleotides $\{\mathtt{A}, \mathtt{C}, \mathtt{G}, \mathtt{U}\}$ and needs to be converted into numerical representations for machine learning.

A widely used conversion technique is using one-hot encoding~\cite{Baldi2001}, which converts the nucleotide in each position of a DNA sequence of length $n_{d}$ into a four-dimensional binary vector and then concatenates each of the $n_{d}$ four-dimensional vectors into a $4n_{d}$-dimensional vector representing the whole sequence.
For instance, let $s \in \mathcal{S}$ where $\mathcal{S} = \{\mathtt{A}, \mathtt{C}, \mathtt{G}, \mathtt{U}\}$, then, a sequence $\mathbf{s} = (\mathtt{A},\mathtt{G},\mathtt{U},\mathtt{U})$ where $n_{d} = 4$ is encoded into a tuple of four 4-dim binary vectors: %\textcolor{\bhlcolor}{$\langle [1,0,0,0],[0,0,1,0],[0,0,0,1],[0,0,0,1] \rangle$}.
\begin{align*}
	\langle [1,0,0,0],[0,0,1,0],[0,0,0,1],[0,0,0,1] \rangle.
\end{align*}

According to recent studies~\cite{lee2015dna,lee2015boosted}, however, applying the vanilla one-hot encoding scheme tends to give limited generalization performance caused by the sparsity of the encoding. In lieu of the one-hot encoding scheme, our approach thus encodes each nucleotide into a four-dimensional dense vector that is randomly initialized and trained by the gradient descent method through the whole architecture as a normal neural network layer~\cite{chollet2015}.
We encode sequences to $\mathbf{x}^{\text{mi}}$ for miRNA and $\mathbf{x}^{\text{m}}$ for mRNA, where $\mathbf{x}$ is a tuple of $n_{d}$ four-dimensional dense vectors.
For instance, a one-hot encoded tuple $\langle [0,0,0,1],[0,0,1,0] \rangle$ is encoded into a dense tuple %$\langle [-0.1,-0.2,0.1,0.3],[-0.1,-0.2,0.3,0.4] \rangle$.
\begin{align*}
	\langle [-0.1,-0.2,0.1,0.3],[-0.1,-0.2,0.3,0.4] \rangle.
\end{align*}

\subsection{Modeling RNAs using RNN based Autoencoder}\label{sec:methods-ae}
Menor \emph{et al.}~\cite{menor2014mirmark} described 151 site-level features between miRNA-target pairs for target prediction, categorizing them into seven groups: binding energy, the type of seed matching, miRNA pairing, target site accessibility, target site composition, target site conservation, and the location of target sites.
To relieve these feature engineering required for target prediction in conventional approaches, \tool~exploits the unsupervised feature learning using the RNN encoder-decoder model~\cite{srivastava2015unsupervised}.

Each of our autoencoder model consists of two RNNs, one as an encoder and the other as a decoder. The encoder RNN encodes $\mathbf{x}$ to $\mathbf{h}$, and the following decoder RNN decodes $\mathbf{h}$ to reconstructed $\mathbf{\hat{x}}$, which minimizes the reconstruction error
\begin{equation}
\mathcal{L}(\mathbf{x}, \mathbf{\hat{x}}) = \lVert \mathbf{x} - \mathbf{\hat{x}} \rVert^{2}
\end{equation}
where $\mathbf{h}$ is a tuple of $n_{d}$ $n_{h}$-dimensional vectors (here $n_{h}$ is the number of the hidden units of the encoder). Through unsupervised learning of these autoencoders, we get the representations of inherent features that will be used by the stacked RNN layer described in Section~\ref{sec:methods-rnn}.

After unsupervised learning of these two autoencoders for miRNA and mRNA, respectively, we bypass the decoder and connect the encoder directly to the next layer as a tuple of the fixed dimension $n_{h}$ representations as shown in \figurename~\ref{fig:methods-overview}.
This restriction of the fixed dimension $n_{h}$ is to model the activations of the second layer to be analogous to target pairing patterns.
We then obtain sequence representations $\mathbf{h}^{\text{mi}}$ for miRNA and $\mathbf{h}^{\text{m}}$ for mRNA.

\subsection{Modeling Interaction between RNAs}\label{sec:methods-rnn}
The miRNA-mRNA binding sites can be classified into three types~\cite{maziere2007prediction}: 5'-dominant canonical, 5'-dominant seed only, and 3'-compensatory. Regardless of the type of a binding site, its detection typically requires the sequence alignment procedure in most of the conventional approaches. However, sequence alignment may often limit the robustness of target prediction because of the need to tune various alignment parameters. The use of an RNN-based approach allows us to omit the sequence alignment procedure (Refer to \figurename~\ref{fig:results-activation} to see that our RNN-based approach can still detect the binding sites without any alignment).

In this interaction modeling step, the learned features of miRNAs $\mathbf{h}^{\textrm{mi}}$ and mRNAs $\mathbf{h}^{\textrm{m}}$ (extracted by the two autoencoders described in Section~\ref{sec:methods-ae}) need to be combined into one tuple $\mathbf{h}$. To this end, we can devise various ways to merge the representations between each model (\eg, summation, multiplication, concatenation, and average).

The concatenated representations from different modalities have been unsuccessful in constructing a high-dimensional vector since the correlation between features in each modality is stronger than that between modalities~\cite{ngiam2011multimodal}. To resolve this issue, deep learning methods have been proposed to learn joint representations that are shared across multiple modalities after learning modality-specific network layers~\cite{sohn2014improved}.
In this paper, we adopt the method of concatenating each dimension, $\mathbf{h}^{\text{mi}}$ and $\mathbf{h}^{\text{m}}$, into tuple $\mathbf{h}$ given by
\begin{align}
	\mathbf{h} = \langle \mathbf{h}_{1}, \cdots, \mathbf{h}_{n_{d}} \rangle = \langle (\mathbf{h}_{1}^{\text{mi}}, \mathbf{h}_{1}^{\text{m}}), \cdots, (\mathbf{h}_{n_{d}}^{\text{mi}}, \mathbf{h}_{n_{d}}^{\text{m}}) \rangle.
\end{align}

The two unsupervised-learned encoders followed by a stacked RNN layer comprises our proposed architecture for target prediction that has $\mathbf{x}^{\text{mi}}$ and $\mathbf{x}^{\text{m}}$ as the inputs.
%This architecture that has two hidden layers of two encoders and one RNN is to learn the interaction between miRNA and mRNA by minimizing the logarithmic loss.
The output node has an activation probability given by
\begin{align}
	P(Y=i | \mathbf{h}) = \frac{1/(1+\mathrm{exp}(-\mathbf{w}_{i}^{T}\mathbf{h}))}{\sum_{k=0}^{1} 1/(1+\mathrm{exp}(-\mathbf{w}_{k}^{T}\mathbf{h}))}	
\end{align}

\noindent where $y$ is the label whether the given pair is a \textcolor{\finalcolor}{miRNA-target pair} ($y=1$) or not ($y=0$), and \textcolor{\finalcolor}{$k\in\{0,1\}$} is the class index. The objective function $\mathcal{L}(\mathbf{w})$ that has to be minimized is then as follows:
\begin{equation}
    \mathcal{L}(\mathbf{w}) = -\frac{1}{N} \sum_{i=1}^{N} (p_{i}\textrm{log}(p_{i}) + (1-p_{i})\textrm{log}(1-p_{i}))
\end{equation}
%subject to
%\begin{equation}
%    \argmax_{y} P(Y=y | \mathbf{x}^{\textrm{mi}}, \mathbf{x}^{\textrm{m}})
%\end{equation}

\noindent where $p_{i}$ is the probability of the given pair \textcolor{\finalcolor}{($\mathbf{x}^{\textrm{mi}}, \mathbf{x}^{\textrm{m}})$ being a valid miRNA-target pair} and $N$ is the mini-batch size used.

\section{Experimental Results}
\begin{figure}
    \centering
    \includegraphics[width=\linewidth]{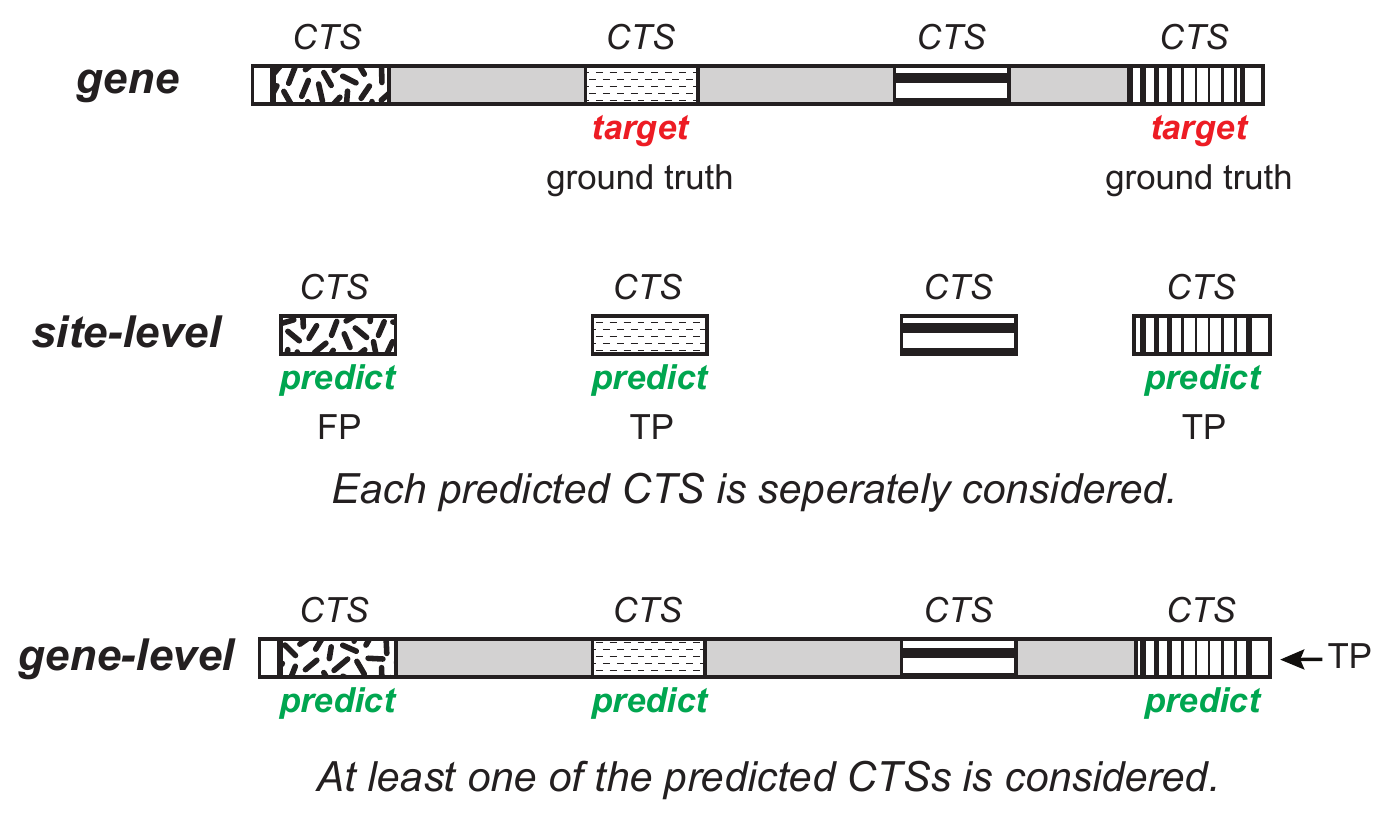}
    \caption{Definition of site-level and gene-level target pair datasets: Assume that there are four CTSs on a genome, and only two of them are true target sites (labeled `target' in the top figure). Suppose that a target prediction tool predicts the first, second, and the fourth CTS as potential targets (labeled `predict' in the middle and bottom figures). A site-level dataset has a label for each CTS, and we can use it to evaluate the tool performance at site level. By contrast, a gene-level dataset has a level for each gene, and we can evaluate the performance of the prediction tool only at gene level.}
    \label{fig:results-def-pair}
\end{figure}

\subsection{Experiment Setup}
\subsubsection{Dataset}
In the literature, there exist two types of datasets that contain miRNA-mRNA pairing information: \emph{site-level} and \emph{gene-level} datasets. As depicted in \figurename~\ref{fig:results-def-pair}, let us assume that there are four CTSs on a gene sequence, two of which are true target sites of a miRNA. A site-level dataset has a label for each CTS indicating whether this CTS is a target or not, and we can evaluate the performance of a target prediction tool for each CTS. On the other hand, a gene-level dataset has a label for each gene, not for each CTS, and we can evaluate the prediction performance at the gene level. Refer to the caption for \figurename~\ref{fig:results-def-pair} for additional details.

To test the proposed \tool, we utilized a public human miRNA-mRNA pairing data repository~\cite{menor2014mirmark}, which provides both site-level and gene-level datasets. The creators of this repository obtained validated target sites from miRecords~\cite{xiao2009mirecords} database and mature human miRNA sequences from mirBase~\cite{griffiths2008mirbase} database. We used 2,042 human miRNAs that are known to bind to their cognate mRNAs.
To train \tool, we utilized the 507 site-level and 2,891 gene-level miRNA-target site pairs as the positive training set. The negative training set was generated as explained in the next subsection.

%While considering miRNA-target site pairs, there are two types of datasets as shown in \figurename~\ref{fig:results-def-pair}: site-level and gene-level target pairs. The site-level pairs are composed by excluding unresolvable target positions from UCSC Genome Browser.
%In this work, the resulting list of 507 for site-level and 2,891 for gene-level miRNA-target site pairs were used as the positive dataset.

\subsubsection{Negative Training Data Generation}
The negative dataset was generated using mock miRNAs in a similar manner to~\cite{john2004human,maragkakis2009accurate} that did not have overlap with the seed sequences of any real mature miRNAs in mirBase. The mock miRNAs were generated with random permutaions using the Fisher-Yates shuffle algorithm~\cite{fisher1949statistical}.
The 507 for site-level and 3,133 for gene-level negative mock miRNA-target pairs were then generated for each real miRNA-target pair in the positive dataset by replacing the positive miRNA in the real miRNA-target pair. The negative target regions were generated for each mock miRNA-target pair using MiRanda~\cite{enright2004microrna} by thresholding the minimum alignment score for minimizing the biases relevant to the miRanda algorithm.

Note that the negative data generation procedure described above is widely used in the literature to handle lack of biologically meaningful and experimentally verified negative pairs for training. For example, the mock miRNA-based negative dataset we used is essentially identical to that used in~\cite{john2004human,maragkakis2009accurate,menor2014mirmark}.
In addition, as mentioned in~\cite{menor2014mirmark}, the mock miRNA-based negative dataset in lieu of real biological negative ones was chosen to obtain a balanced training dataset.
These authors investigated four combinations of training and test datasets with the real and mock sequences and found that the mock miRNAs for both model building and validation had the best predictive performance.

\begin{figure*}[!t]
	\centering
    \begin{minipage}[b]{.475\textwidth}
        \centering
        \includegraphics[width=.9\textwidth]{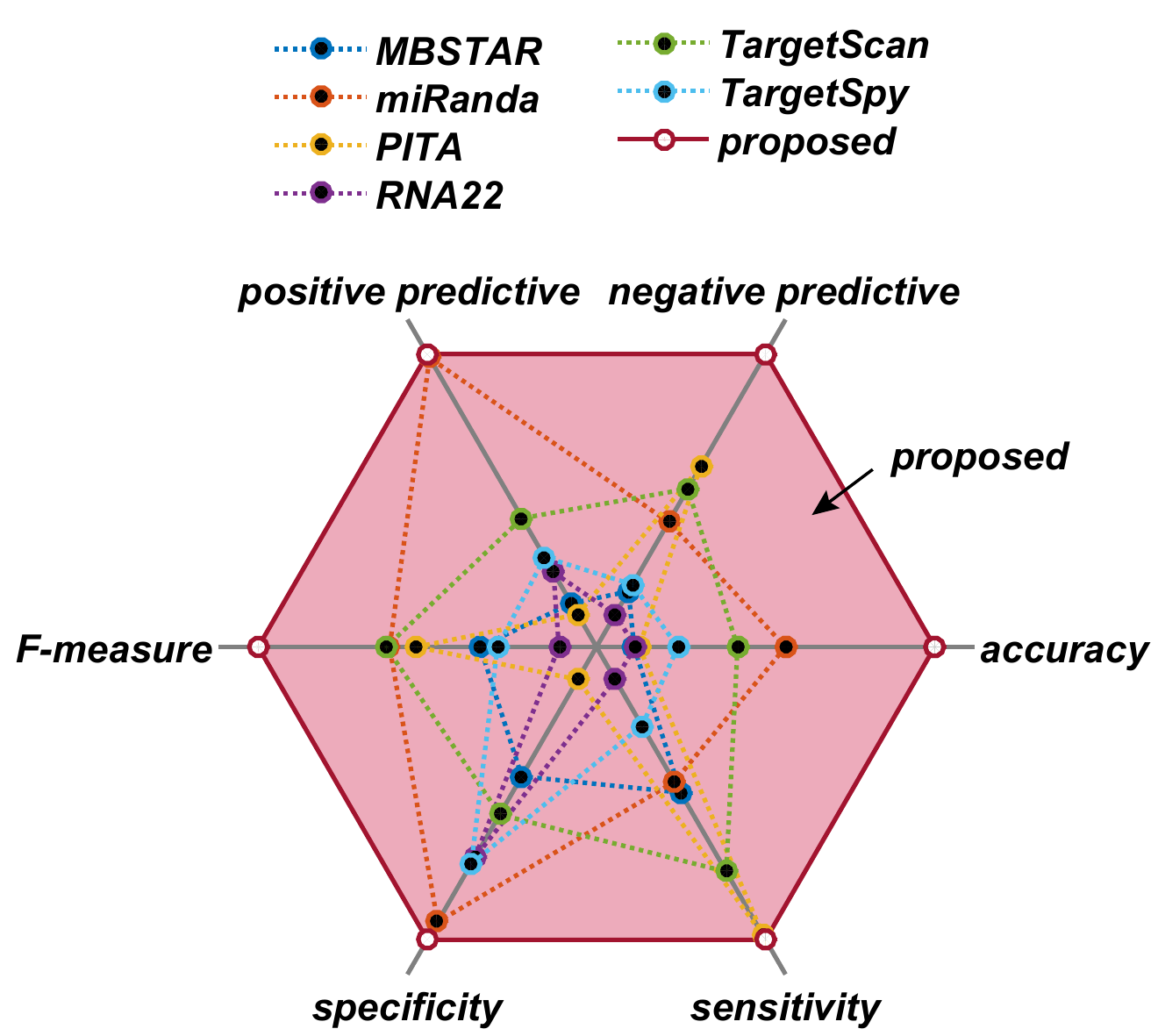}
        \caption{Comparison of prediction performance of proposed \tool~and alternatives (best viewed in color). The proposed \tool~outperforms the compared alternatives in terms of all the evaluation metrics used (\textcolor{\bhlcolor}{F-measure}, specificity, sensitivity, accuracy, negative predictive value, and positive predictive value; see the footnote on \textcolor{\bhlcolor}{page 5} for the definitions of these metrics \textcolor{\bhlcolor}{and Table~\ref{tbl:results-accuracy-tool} for the values of each metric})}
        \label{fig:results-accuracy-tool}
    \end{minipage}
    \quad
    \begin{minipage}[b]{.475\textwidth}
        \centering
        \includegraphics[width=.9\textwidth]{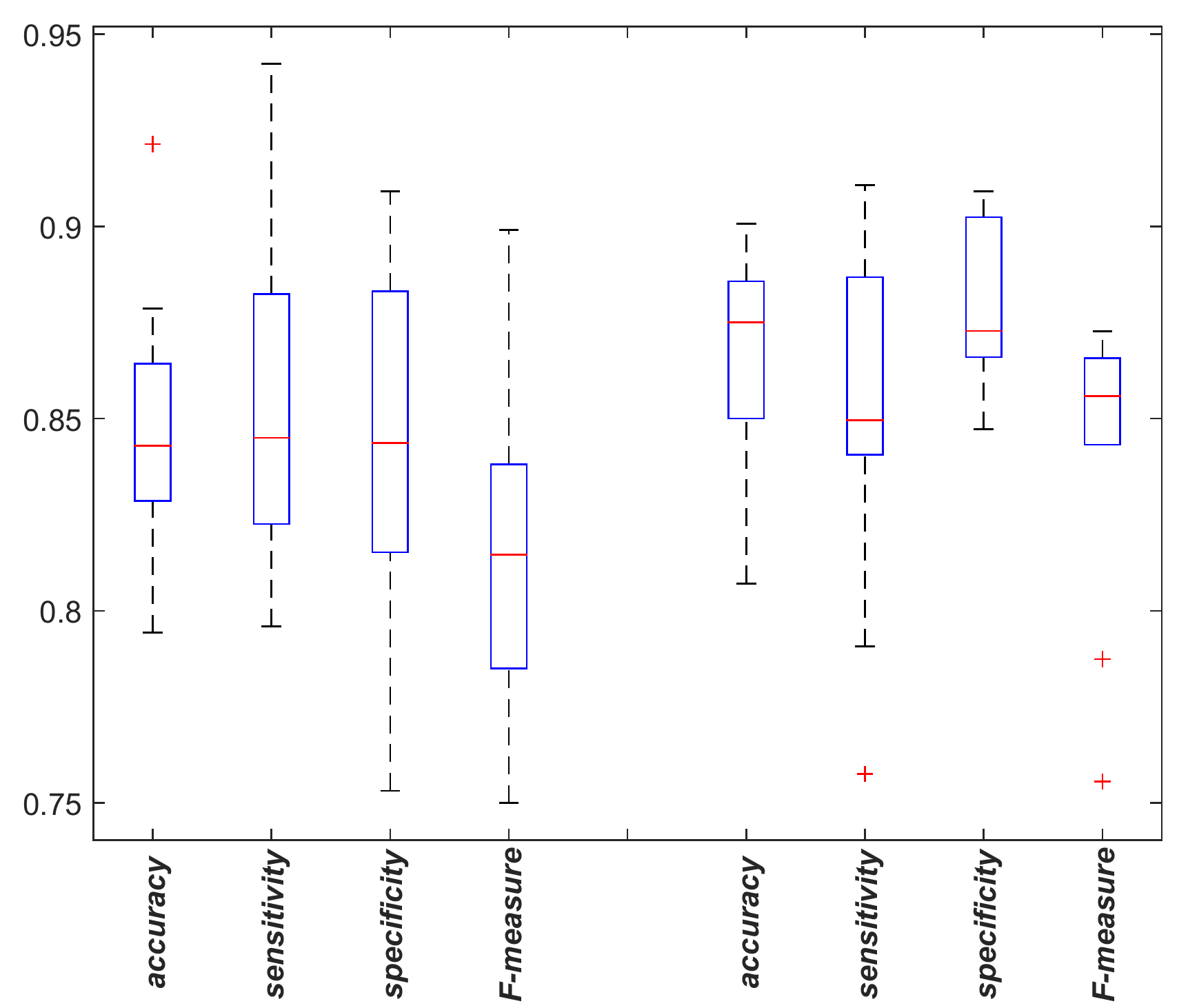}
        \caption{Test accuracy of our approach using two types of memory units (left: deep RNN with LSTM units, right: with GRU units)}
        \label{fig:results-accuracy-fold}
    \end{minipage}
\end{figure*}

\begin{spacing}{\mytablespacing}
\setlength{\tabcolsep}{15pt}
\ctable[
    caption = {Comparison of prediction performance of proposed \tool~and alternatives},
    label = {tbl:results-accuracy-tool},
    doinside = \small,
    width = 2\columnwidth,
    star
]
{crrrrrrr}
{}
{
    \toprule
          & \multicolumn{1}{c}{MBSTAR} & \multicolumn{1}{c}{miRanda} & \multicolumn{1}{c}{PITA} & \multicolumn{1}{c}{RNA22} & \multicolumn{1}{c}{TargetScan} & \multicolumn{1}{c}{TargetSpy} & \multicolumn{1}{c}{deepTarget} \\
    \midrule
    accuracy & 0.5805 & 0.7737 & 0.5870 & 0.5820 & 0.7125 & 0.6364 & \textbf{0.9641} \\
    sensitivity & 0.5308 & 0.6122 & 0.9301 & 0.4006 & 0.7980 & 0.4934 & \textbf{0.9389} \\
    specificity & 0.5308 & 0.9228 & 0.2703 & 0.7494 & 0.6336 & 0.7657 & \textbf{0.9706} \\
    F-measure & 0.5921 & 0.7220 & 0.6837 & 0.4791 & 0.7271 & 0.5672 & \textbf{0.9105} \\
    PPV   & 0.5551 & 0.8797 & 0.5405 & 0.5960 & 0.6677 & 0.6616 & \textbf{0.8848} \\
    NPV   & 0.6114 & 0.7206 & 0.8074 & 0.5753 & 0.7727 & 0.6223 & \textbf{0.9845} \\
    \bottomrule
}
\end{spacing}

\subsubsection{Network Architecure and Training}
In the following sections, the architecture of our method is denoted by the number of nodes in each layer. For example, an architecture 4-60-30-2 has four layers with four input units, 60 units in the first hidden layer, and so on.

The proposed RNN-based approach used [(4-30-4) $\parallel$ (4-30-4)]-(60-30)-2 where (4-30-4) is for the autoencoder described in Section~\ref{sec:methods-ae} and (60-30) is for the stacked RNN described in Secion~\ref{sec:methods-rnn}. Note that the symbol `$\parallel$' represents the two parallel autoencoders.
The number of output units of each encoder for miRNA and mRNA, respectively, was set to an equal number in order to model the activations of the second layer to be analogous to target pairing patterns.

For training, our approach optimized the logarithmic loss function using Adam~\cite{kingma2014adam} [batch size: 50, the number of epochs: 50 (pre-training autoencoder), and 400 (fine-tuning architecture)]. The weights were initialized according to a uniform distribution as directed in~\cite{glorot2010understanding}. We used dropout as the regularizer in lieu of recently popularized batch normalization~\cite{ioffe2015batch} with 1,014 site-level target pairs.

The effects of varying architectures and dropout parameters will be presented in Section 4.3.

\subsection{Prediction Performance}\label{sec:results-performance}
To evaluate the prediction performance of our \tool~approach, we compared its performance with six existing tools for miRNA target prediction: MBSTAR~\cite{bandyopadhyay2015mbstar}, miRanda, PITA~\cite{kertesz2007role}, RNA22~\cite{miranda2006pattern}, TargetScan~\cite{lewis2003prediction}, and TargetSpy~\cite{sturm2010targetspy} with 6,024 gene-level target pairs.
\textcolor{\bhlcolor}{After we found CTSs in 6,024 gene-level target pairs, there remained 4,735 positive and 1,225 negative pairs because of the disappearance of the seed in the mock miRNAs that match the mRNA sequences.}

As a performance metric, we used the widely used measures including accuracy\footnote{$\textrm{accuracy} = (TP+TN) / (TP+TN+FP+FN)$, where $TP$, $FP$, $FN$, and $TN$ represent the numbers of true positives, false positives, false negatives, and true negatives, respectively.}, sensitivity\footnote{$\textrm{sensitivity} = TP / (TP+FN)$}, specificity\footnote{$\textrm{specificity} = TN / (TN+FP)$}, \textcolor{\bhlcolor}{F-measure}\footnote{$\textrm{F-measure} = 2TP /(2TP+FP+FN)$}, positive predictive value (PPV)\footnote{$\textrm{PPV} = TP / (TP+FP)$}, and negative predictive value (NPV)\footnote{$\textrm{NPV} = TN / (TN + FN)$}.

According to the definition of these metrics, they generally have FN in the denominator. Namely, high sensitivity and NPV represent that FN is low; however, the algorithms that classify most of the samples as positive would also have low FN. To distinguish such poor algorithms from more accurate ones, we also have to consider the specificity and PPV metrics, whose definitions include FP in the numerator.

\figurename~\ref{fig:results-accuracy-tool} \textcolor{\bhlcolor}{and Table~\ref{tbl:results-accuracy-tool} show} the comparison of prediction performance.
To obtain the performance of \tool, we performed 10-fold cross validation and averaged each resulting value.
Overall, our method significantly outperformed the best alternative in terms of accuracy, sensitivity, and \textcolor{\bhlcolor}{F-measure by 24.61\%, 17.66\%, and 25\%}, respectively. As will be discussed in Section 5, biologists tend to give more priority to sensitivity than to specificity, when searching for potential targets of specific miRNA~\cite{m2011practical}. %We can thus interpret that our approach has high predictive performance than the alternatives.
%According to~\cite{m2011practical}, biologists give priority to sensitivity in the search for potential targets of specific miRNA and to specificity in the examination of miRNAs that regulate specific gene, hence, we can interpret that our approach has high predictive performance than the alternatives.

%\subsection{Effects of Architecture and Hyperparameter Variation on Performance}
\subsection{Effects of Architecture Variation}
\figurename~\ref{fig:results-accuracy-fold} shows the effects of the memory unit used in \tool~on its prediction performance. To generate this plot, we performed 10-fold cross validation and tested two popular memory units, LSTM and GRU.
According to Jozefowicz \etal~\cite{jozefowicz2015empirical}, it is believed that there is no architecture that can consistently beat the LSTM and the GRU in all conditions. %, hence, we adopt the two memory units, LSTM and GRU.

As shown in \figurename~\ref{fig:results-accuracy-fold}, the architecture composed of GRU outperforms that of LSTM in our experiments. More quantitatively, the architecture using GRU as its memory unit showed 3.81\%, 0.53\%, 3.46\%, and 5.06\% improvements over the LSTM-installed architecture in terms of accuracy, sensitivity, specificity, and \textcolor{\bhlcolor}{F-measure}, respectively (the median values achieved were 0.8750, 0.8496, 0.8728, and 0.8559, respectively, in 10-fold cross validation).

Table~\ref{tbl:results-architecture} lists the effect of varying the architecture of the second layer RNN on overall performance. We tested stacked RNNs with one to three layers and single-layer bidirectional RNNs. Using the two layer architecture gave the best overall results, although the differences in performance were not significant.

%As we increased the number of the stacked RNNs, we could not observe any significant improvement. %This result suggess that there is only a little meaningful hierarchical features for target prediction.

%\subsection{Tuning Hyperparameters}
Table~\ref{tbl:results-dropout} presents the effect of changing the probability of dropout on performance. The architecture with dropping hidden unit activations with a probability of 0.1 during training gave the best performance, delivering 0.61--3.44\% improvements over the other probability values tried.

\begin{spacing}{\mytablespacing}
\setlength{\tabcolsep}{10pt}
\ctable[
    caption = {Effects of architecture variation on prediction performance of \tool},
    label = {tbl:results-architecture},
    doinside = \small,
    width = \columnwidth
]
{ccccc}
{}
{
    \toprule
          \multicolumn{1}{c}{Architecture} & \multicolumn{1}{c}{1-layer} & \multicolumn{1}{c}{2-layer} & \multicolumn{1}{c}{3-layer} & \multicolumn{1}{c}{bidirectional} \\
    \midrule
    accuracy & 0.8214 & \textbf{0.8286} & 0.8250 & 0.8250 \\
    sensitivity & \textbf{0.8429} & 0.8351 & 0.8417 & 0.8254 \\
    specificity & 0.8204 & \textbf{0.8428} & 0.8297 & 0.8041 \\
    F-measure & 0.7745 & \textbf{0.7838} & 0.7759 & 0.7620 \\
    \bottomrule
}
\end{spacing} 
\begin{spacing}{\mytablespacing}
\setlength{\tabcolsep}{8pt}
\ctable[
    caption = {Effects of changing dropout probability on prediction performance of \tool},
    label = {tbl:results-dropout},
    doinside = \small,
    width = \columnwidth
]
{cccccc}
{}
{
    \toprule
          \multicolumn{1}{c}{Dropout} & \multicolumn{1}{c}{0.0} & \multicolumn{1}{c}{0.1} & \multicolumn{1}{c}{0.2} & \multicolumn{1}{c}{0.3} & \multicolumn{1}{c}{0.4} \\
    \midrule
    accuracy & 0.8214 & \textbf{0.8429} & 0.8179 & 0.8179 & 0.7857 \\
    sensitivity & 0.8429 & \textbf{0.8480} & 0.8460 & 0.8460 & 0.7980 \\
    specificity & 0.8204 & \textbf{0.8486} & 0.8111 & 0.8111 & 0.7746 \\
    F-measure & 0.7745 & \textbf{0.7978} & 0.7592 & 0.7592 & 0.7386 \\
    \bottomrule
}
\end{spacing} 
%\input{tbl-results-cts-size}

%\begin{figure*}
%	\centering
%    \begin{minipage}[b]{.45\textwidth}
%        \centering
%        \includegraphics[width=\linewidth]{results-distribution}
%        \caption{Distribution of the length of the target sites}
%        \label{fig:results-distribution}
%    \end{minipage}
%    \quad
%    \begin{minipage}[b]{.5\textwidth}
%        \centering
%        \includegraphics[width=.9\linewidth]{results-activation}
%        \caption{Change of activation of each node in the RNN layer over RNA sequences [darker shades mean higher activation values; the red box indicates the activation sequence of the hidden unit that best represents the binding site pattern]}
%        \label{fig:results-activation}
%    \end{minipage}
%\end{figure*}

%\figurename~\ref{fig:results-distribution} and Table~\ref{tbl:results-cts-size} are to show how we determined the  $k$ parameter (the size of CTS; see Section~2.3). As depicted in \figurename~\ref{fig:results-distribution}, 51\% of the real target regions were of a length of 21 or 22, which re-confirms the biological fact that miRNAs usually contains approximately 22 nucleotides each. In our experiments, we tested various CTS sizes, varying $k$ from 10 to 50 by units of 5, and the resulting distribution is shown in \figurename~\ref{fig:results-distribution}. As listed in Table~\ref{tbl:results-cts-size}, the CTS of size 40 gave the best predictive performance in our experiments, and we thus used $k=40$ in the preprocessing step for handling the CTS.

\subsection{Visual Inspection of RNN Activations}
\begin{figure}
    \centering
    \includegraphics[width=.9\linewidth]{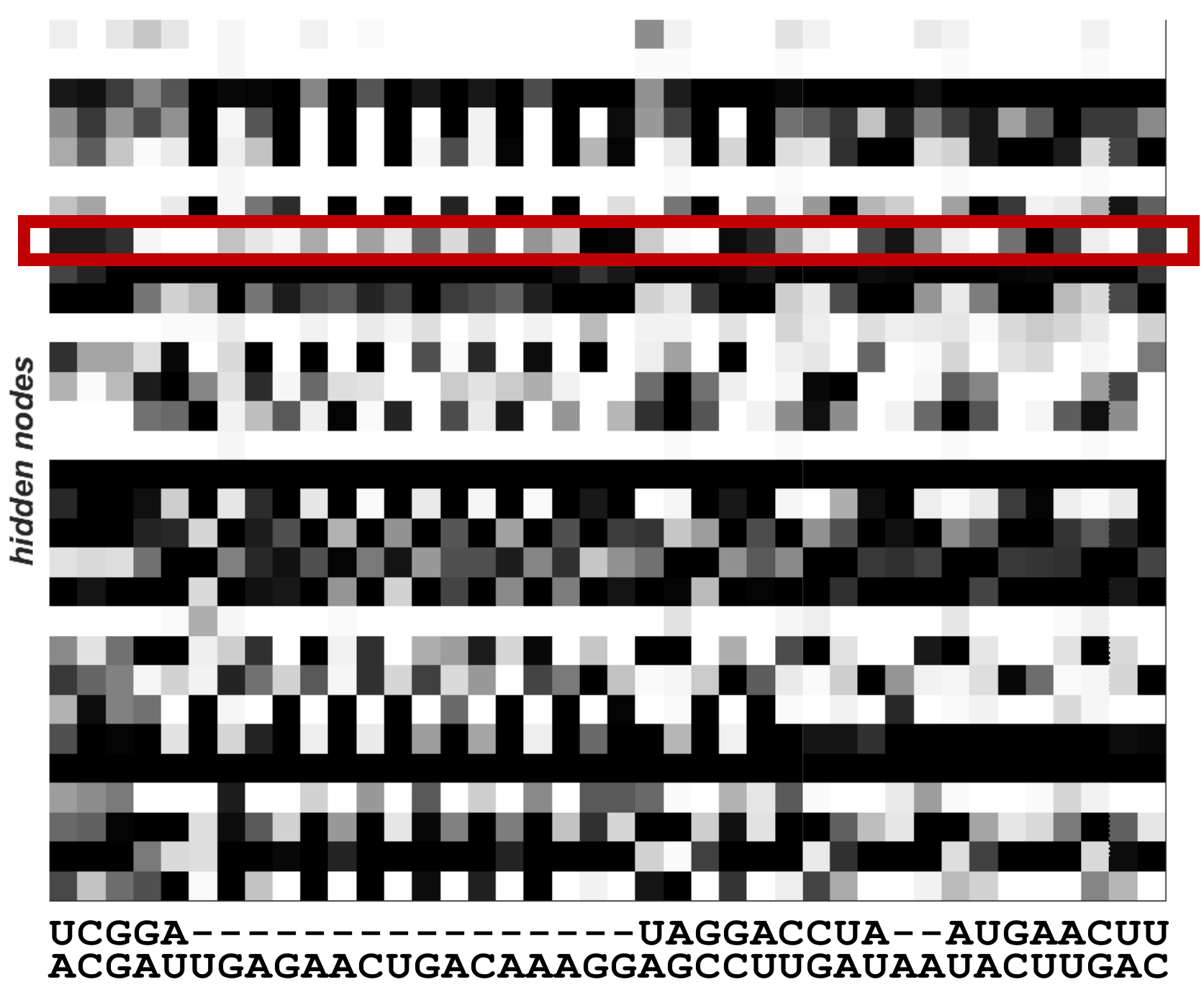}
    \caption{Change of activation of each node in the RNN layer over RNA sequences [darker shades mean higher activation values; the red box indicates the activation sequence of the hidden unit that best represents the binding site pattern]}
    \label{fig:results-activation}
\end{figure}

\figurename~\ref{fig:results-activation} shows the activations in the RNN layer in the second stage for learning sequence-to-sequence interactions. The $x$-axis represents two RNA sequences (top: miRNA, bottom: mRNA), and the $y$-axis represents the position of each hidden unit in the layer. Using this plot, we can see how the activation changes over the nucleotide positions along the RNA sequences. Note that darker shades represents higher activation values. The red box indicates the activation sequence of the hidden unit that best represents the binding site pattern. This pattern seems compatible with the sequence alignment results shown by the two aligned RNA sequences in the $x$-axis. These results suggest that analyzing the activation patterns may allow us to gain novel biological insight into regulatory interactions.

%As depicted in \figurename~\ref{fig:results-activation}, for instance, we observed that the series of the activation of the second hidden unit (red boxed row) detects the complementary pairs of binding without sequence alignment operation. 

\section{Discussion}
Biological sequences (such as DNA, RNA, and protein sequences) naturally fit the recurrent neural networks that are capable of temporal modeling. Nonetheless, prior work on applying deep learning to bioinformatics utilized only convolutional and fully connected neural networks. The biggest novelty of our work lies in its use of recurrent neural networks to model RNA sequences and further learn their sequence-to-sequence interactions, without laborious feature engineering (\eg, more than 151 features of miRNA-target pairs have been proposed in the literature). As shown in our experimental results, even without any of the known features, \tool~delivered substantial performance boosts (over 25\% increase in F-measure) over existing miRNA target detectors, demonstrating the effectiveness of recent advances in end-to-end learning methodologies.

\textcolor{\bhlcolor}{When dealing with genome-wide applications, such as miRNA target prediction, we often need to handle extremely imbalanced datasets in which negative examples significantly outnumber positive examples. According to Saito and Rehmsmeier~\cite{saito2015precision}, as PPV changes with the ratio of positives and negatives, PPV is more useful than the other metrics for evaluating binary classifiers on imbalanced datasets. As mentioned in Section~\ref{sec:results-performance}, \tool~was trained with 4,735 positive and 1,225 negative pairs and showed higher performance than the alternatives in terms of PPV. The higher values of sensitivity and PPV deepTarget reported from imbalance datasets indicate that deepTarget performs more robust prediction of miRNA-mRNA pairs.}

When training \tool, we focused on improving its capability to reject false positives (\ie, bogus miRNA-mRNA pairs) as a target predictor. Given that there is a trade-off between sensitivity and specificity, our training principle resulted in slight loss in the specificity of \tool. This decision was based on the study that more priority should be given to sensitivity in the search for potential targets of specific miRNAs, whereas specificity should be emphasized in the examination of miRNAs that regulate specific genes~\cite{m2011practical}. Depending on the specific needs, we could alternatively train \tool~to put more priority on specificity.
For instance, this could be done by altering the composition of a mock negative dataset to have additional mispairings between miRNA and mRNA sequences except the seed sequence.

Notably, \tool~does not depend on any sequence alignment operation, which has been used in many bioinformatics pipelines as a holy grail to reveal similarity/interactions between sequences. Although effective in general, sequence alignment is susceptible to changes in parameters (\eg, gap/mismatch penalty and match premium) and often fails to reveal the true interactions between sequences, as is often observed in most of the alignment-based miRNA target detectors. By processing miRNA and RNA sequences with RNN-based auto-encoders without alignment, \tool~successfully discover the inherent sequence representations, which are effectively used in the next step of \tool~for interaction learning.

In Figure~\ref{fig:results-activation}, we visualized the activations in the RNN layers of \tool~as an attempt to discriminate between the positive and negative samples.
Contrasting and scrutinizing the patterns existing in each class will be helpful for providing insight that can eventually lead to discovery of novel biological features.
Given that biomedical practitioners prefer `white-box' approaches (\eg, logistic regression) to `black-box' approaches (\eg, deep learning), such efforts will hopefully expedite the deployment and acceptance of deep learning for biomedicine~\cite{min2016deep}.
In this regard, attempts to decode the information learned by deep learning are emerging (\eg, we can compute importance scores for each activation to better interpret activation energies~\cite{shrikumar2016not}).
%may be adapted to better interpret activation energies to gain deeper biological insights.

Although the performance of \tool~is incomparably higher than that of the existing tools compared, there still remains room for further improvements. An additional breakthrough may be possible by enhancing the current step to learn sequence-to-sequence interactions. The current version of \tool~relies on concatenating the RNA representations from two auto-encoders and learning interactions therein using a unidirectional two-layer RNN architecture. Although this architecture was effective to some extents, as shown in our experiments, adopting even more sophisticated approaches may further boost the capability of \tool~to detect subtle interactions that currently go undetected. Our future work thus includes investigating how to improve the interaction learning part, which is crucial for achieving additional performance gains.

%\textcolor{red}{(I omitted comments on batch normalization; please mention that we did not use it and instead used dropout somewhere in the manuscript.)}

%Robust \emph{in silico} prediction of miRNA targets has been challenging, and numerous computational approaches have been proposed. However, with limited success. Th , mainly due to the small alphabet size (\ie, only four letters: \texttt{A}, \texttt{C}, \texttt{G}, \texttt{T}), 

\section*{Acknowledgment}
This work was supported in part by the National Research Foundation of Korea (NRF) grant funded by the Korea government (Ministry of Science, ICT and Future Planning) [No. 2012M3A9D1054622, No. 2014M3C9A3063541, No. 2014M3A9E2064434, and No. 2015M3A9A7029735], and in part by the Brain Korea 21 Plus Project in 2016.

\balance
\bibliographystyle{abbrv}
\bibliography{references}

\begin{thebibliography}{10}

\bibitem{bahdanau2014neural}
D.~Bahdanau, K.~Cho, and Y.~Bengio.
\newblock Neural machine translation by jointly learning to align and
  translate.
\newblock {\em arXiv preprint arXiv:1409.0473}, 2014.

\bibitem{Baldi2001}
P.~Baldi and S.~Brunak.
\newblock Chapter 6. neural networks: applications.
\newblock In {\em Bioinformatics: The Machine Learning Approach}. MIT press,
  2001.

\bibitem{bandyopadhyay2015mbstar}
S.~Bandyopadhyay, D.~Ghosh, R.~Mitra, and Z.~Zhao.
\newblock {MBSTAR: multiple instance learning for predicting specific
  functional binding sites in microRNA targets}.
\newblock {\em Scientific reports}, 5, 2015.

\bibitem{bartel2004micrornas}
D.~P. Bartel.
\newblock {MicroRNAs}: genomics, biogenesis, mechanism, and function.
\newblock {\em Cell}, 116(2):281--297, 2004.

\bibitem{chengmirtdl}
S.~Cheng, M.~Guo, C.~Wang, X.~Liu, Y.~Liu, and X.~Wu.
\newblock {MiRTDL: a deep learning approach for miRNA target prediction}.

\bibitem{cho2014properties}
K.~Cho, B.~van Merri{\"e}nboer, D.~Bahdanau, and Y.~Bengio.
\newblock {On the properties of neural machine translation: Encoder-decoder
  approaches}.
\newblock {\em arXiv preprint arXiv:1409.1259}, 2014.

\bibitem{cho2014learning}
K.~Cho, B.~Van~Merri{\"e}nboer, C.~Gulcehre, D.~Bahdanau, F.~Bougares,
  H.~Schwenk, and Y.~Bengio.
\newblock Learning phrase representations using {RNN} encoder-decoder for
  statistical machine translation.
\newblock {\em arXiv preprint arXiv:1406.1078}, 2014.

\bibitem{chollet2015}
F.~Chollet.
\newblock {Keras: Deep Learning library for Theano and TensorFlow}.
\newblock \url{https://github.com/fchollet/keras}, 2015.

\bibitem{enright2004microrna}
A.~J. Enright, B.~John, U.~Gaul, T.~Tuschl, C.~Sander, D.~S. Marks, et~al.
\newblock {MicroRNA targets in Drosophila}.
\newblock {\em Genome Biology}, 5(1):R1--R1, 2004.

\bibitem{fisher1949statistical}
R.~A. Fisher, F.~Yates, et~al.
\newblock Statistical tables for biological, agricultural and medical research.
\newblock {\em Statistical tables for biological, agricultural and medical
  research.}, (Ed. 3.), 1949.

\bibitem{friedman2009most}
R.~C. Friedman, K.~K.-H. Farh, C.~B. Burge, and D.~P. Bartel.
\newblock {Most mammalian mRNAs are conserved targets of microRNAs}.
\newblock {\em Genome Research}, 19(1):92--105, 2009.

\bibitem{glorot2010understanding}
X.~Glorot and Y.~Bengio.
\newblock Understanding the difficulty of training deep feedforward neural
  networks.
\newblock In {\em International conference on artificial intelligence and
  statistics}, pages 249--256, 2010.

\bibitem{Goodfellow-et-al-2016-Book}
I.~Goodfellow, Y.~Bengio, and A.~Courville.
\newblock Deep learning.
\newblock Book in preparation for MIT Press, 2016.

\bibitem{gregor2015draw}
K.~Gregor, I.~Danihelka, A.~Graves, and D.~Wierstra.
\newblock {DRAW}: A recurrent neural network for image generation.
\newblock {\em arXiv preprint arXiv:1502.04623}, 2015.

\bibitem{griffiths2008mirbase}
S.~Griffiths-Jones, H.~K. Saini, S.~van Dongen, and A.~J. Enright.
\newblock {miRBase: tools for microRNA genomics}.
\newblock {\em Nucleic Acids Research}, 36(suppl 1):D154--D158, 2008.

\bibitem{hochreiter1997long}
S.~Hochreiter and J.~Schmidhuber.
\newblock Long short-term memory.
\newblock {\em Neural Computation}, 9(8):1735--1780, 1997.

\bibitem{ioffe2015batch}
S.~Ioffe and C.~Szegedy.
\newblock Batch normalization: Accelerating deep network training by reducing
  internal covariate shift.
\newblock {\em arXiv preprint arXiv:1502.03167}, 2015.

\bibitem{john2004human}
B.~John, A.~J. Enright, A.~Aravin, T.~Tuschl, C.~Sander, D.~S. Marks, et~al.
\newblock {Human microRNA targets}.
\newblock {\em PLoS Biol}, 2(11):e363, 2004.

\bibitem{jozefowicz2015empirical}
R.~Jozefowicz, W.~Zaremba, and I.~Sutskever.
\newblock An empirical exploration of recurrent network architectures.
\newblock In {\em Proceedings of the 32nd International Conference on Machine
  Learning (ICML-15)}, pages 2342--2350, 2015.

\bibitem{kertesz2007role}
M.~Kertesz, N.~Iovino, U.~Unnerstall, U.~Gaul, and E.~Segal.
\newblock {The role of site accessibility in microRNA target recognition}.
\newblock {\em Nature Genetics}, 39(10):1278--1284, 2007.

\bibitem{kingma2014adam}
D.~Kingma and J.~Ba.
\newblock Adam: A method for stochastic optimization.
\newblock {\em arXiv preprint arXiv:1412.6980}, 2014.

\bibitem{kingma2013auto}
D.~P. Kingma and M.~Welling.
\newblock Auto-encoding variational {Bayes}.
\newblock {\em arXiv preprint arXiv:1312.6114}, 2013.

\bibitem{le2015simple}
Q.~V. Le, N.~Jaitly, and G.~E. Hinton.
\newblock A simple way to initialize recurrent networks of rectified linear
  units.
\newblock {\em arXiv preprint arXiv:1504.00941}, 2015.

\bibitem{lee2015dna}
B.~Lee, T.~Lee, B.~Na, and S.~Yoon.
\newblock {DNA}-level splice junction prediction using deep recurrent neural
  networks.
\newblock {\em arXiv preprint arXiv:1512.05135}, 2015.

\bibitem{lee2015boosted}
T.~Lee and S.~Yoon.
\newblock Boosted categorical restricted boltzmann machine for computational
  prediction of splice junctions.
\newblock In {\em ICML}, pages 2483–--2492, 2015.

\bibitem{lewis2003prediction}
B.~P. Lewis, I.-h. Shih, M.~W. Jones-Rhoades, D.~P. Bartel, and C.~B. Burge.
\newblock Prediction of mammalian {microRNA} targets.
\newblock {\em Cell}, 115(7):787--798, 2003.

\bibitem{lu2005microrna}
J.~Lu, G.~Getz, E.~A. Miska, E.~Alvarez-Saavedra, J.~Lamb, D.~Peck,
  A.~Sweet-Cordero, B.~L. Ebert, R.~H. Mak, A.~A. Ferrando, et~al.
\newblock {MicroRNA expression profiles classify human cancers}.
\newblock {\em nature}, 435(7043):834--838, 2005.

\bibitem{m2011practical}
T.~M~Witkos, E.~Koscianska, and W.~J~Krzyzosiak.
\newblock Practical aspects of {microRNA} target prediction.
\newblock {\em Current molecular medicine}, 11(2):93--109, 2011.

\bibitem{maragkakis2009accurate}
M.~Maragkakis, P.~Alexiou, G.~L. Papadopoulos, M.~Reczko, T.~Dalamagas,
  G.~Giannopoulos, G.~Goumas, E.~Koukis, K.~Kourtis, V.~A. Simossis, et~al.
\newblock {Accurate microRNA target prediction correlates with protein
  repression levels}.
\newblock {\em BMC Bioinformatics}, 10(1):295, 2009.

\bibitem{maziere2007prediction}
P.~Maziere and A.~J. Enright.
\newblock {Prediction of microRNA targets}.
\newblock {\em Drug discovery today}, 12(11):452--458, 2007.

\bibitem{menor2014mirmark}
M.~Menor, T.~Ching, X.~Zhu, D.~Garmire, and L.~X. Garmire.
\newblock {mirMark: a site-level and UTR-level classifier for miRNA target
  prediction}.
\newblock {\em Genome biology}, 15(10):500, 2014.

\bibitem{min2010got}
H.~Min and S.~Yoon.
\newblock Got target?: Computational methods for {microRNA} target prediction
  and their extension.
\newblock {\em Experimental \& Molecular Medicine}, 42(4):233--244, 2010.

\bibitem{min2016deep}
S.~Min, B.~Lee, and S.~Yoon.
\newblock Deep learning in bioinformatics.
\newblock {\em {Briefings in Bioinformatics}}, in press, 2016.

\bibitem{miranda2006pattern}
K.~C. Miranda, T.~Huynh, Y.~Tay, Y.-S. Ang, W.-L. Tam, A.~M. Thomson, B.~Lim,
  and I.~Rigoutsos.
\newblock A pattern-based method for the identification of {microRNA} binding
  sites and their corresponding heteroduplexes.
\newblock {\em Cell}, 126(6):1203--1217, 2006.

\bibitem{ngiam2011multimodal}
J.~Ngiam, A.~Khosla, M.~Kim, J.~Nam, H.~Lee, and A.~Y. Ng.
\newblock Multimodal deep learning.
\newblock In {\em Proceedings of the 28th international conference on machine
  learning (ICML-11)}, pages 689--696, 2011.

\bibitem{pascanu2013construct}
R.~Pascanu, C.~Gulcehre, K.~Cho, and Y.~Bengio.
\newblock How to construct deep recurrent neural networks.
\newblock {\em arXiv preprint arXiv:1312.6026}, 2013.

\bibitem{peterson2014common}
S.~M. Peterson, J.~A. Thompson, M.~L. Ufkin, P.~Sathyanarayana, L.~Liaw, and
  C.~B. Congdon.
\newblock Common features of {microRNA target} prediction tools.
\newblock {\em Front Genet}, 5:23, 2014.

\bibitem{saito2015precision}
T.~Saito and M.~Rehmsmeier.
\newblock {The Precision-Recall Plot Is More Informative than the ROC Plot When
  Evaluating Binary Classifiers on Imbalanced Datasets}.
\newblock {\em PloS One}, 10(3):e0118432, 2015.

\bibitem{sak2014long}
H.~Sak, A.~Senior, and F.~Beaufays.
\newblock Long short-term memory recurrent neural network architectures for
  large scale acoustic modeling.
\newblock In {\em Proceedings of the Annual Conference of International Speech
  Communication Association (INTERSPEECH)}, 2014.

\bibitem{schuster1997bidirectional}
M.~Schuster and K.~K. Paliwal.
\newblock Bidirectional recurrent neural networks.
\newblock {\em Signal Processing, IEEE Transactions on}, 45(11):2673--2681,
  1997.

\bibitem{shrikumar2016not}
A.~Shrikumar, P.~Greenside, A.~Shcherbina, and A.~Kundaje.
\newblock Not just a black box: Learning important features through propagating
  activation differences.
\newblock {\em arXiv preprint arXiv:1605.01713}, 2016.

\bibitem{sohn2014improved}
K.~Sohn, W.~Shang, and H.~Lee.
\newblock Improved multimodal deep learning with variation of information.
\newblock In {\em Advances in Neural Information Processing Systems}, pages
  2141--2149, 2014.

\bibitem{srivastava2014dropout}
N.~Srivastava, G.~Hinton, A.~Krizhevsky, I.~Sutskever, and R.~Salakhutdinov.
\newblock Dropout: A simple way to prevent neural networks from overfitting.
\newblock {\em The Journal of Machine Learning Research}, 15(1):1929--1958,
  2014.

\bibitem{srivastava2015unsupervised}
N.~Srivastava, E.~Mansimov, and R.~Salakhutdinov.
\newblock Unsupervised learning of video representations using {LSTM}s.
\newblock {\em arXiv preprint arXiv:1502.04681}, 2015.

\bibitem{srivastava2014comparison}
P.~K. Srivastava, T.~R. Moturu, P.~Pandey, I.~T. Baldwin, and S.~P. Pandey.
\newblock A comparison of performance of plant {miRNA} target prediction tools
  and the characterization of features for genome-wide target prediction.
\newblock {\em BMC Genomics}, 15(1):1, 2014.

\bibitem{sturm2010targetspy}
M.~Sturm, M.~Hackenberg, D.~Langenberger, and D.~Frishman.
\newblock {TargetSpy: a supervised machine learning approach for microRNA
  target prediction}.
\newblock {\em BMC Bioinformatics}, 11(1):292, 2010.

\bibitem{vincent2008extracting}
P.~Vincent, H.~Larochelle, Y.~Bengio, and P.-A. Manzagol.
\newblock Extracting and composing robust features with denoising autoencoders.
\newblock In {\em Proceedings of the 25th international conference on Machine
  learning}, pages 1096--1103. ACM, 2008.

\bibitem{vinyals2014show}
O.~Vinyals, A.~Toshev, S.~Bengio, and D.~Erhan.
\newblock Show and tell: A neural image caption generator.
\newblock {\em arXiv preprint arXiv:1411.4555}, 2014.

\bibitem{xiao2009mirecords}
F.~Xiao, Z.~Zuo, G.~Cai, S.~Kang, X.~Gao, and T.~Li.
\newblock {miRecords: an integrated resource for microRNA--target
  interactions}.
\newblock {\em Nucleic Acids Research}, 37(suppl 1):D105--D110, 2009.

\bibitem{yoon2006computational}
S.~Yoon and G.~De~Micheli.
\newblock Computational identification of {microRNAs} and their targets.
\newblock {\em {Birth Defects Research Part C: Embryo Today: Reviews}},
  78(2):118--128, 2006.

\end{thebibliography}

\end{document}